\documentclass{ieeeaccess}
\usepackage{cite}
\usepackage{amsmath,amssymb,amsfonts}
\usepackage{graphicx}
\usepackage{algorithm}
\usepackage{algpseudocode}
\usepackage{bm}
\usepackage{textcomp}

\DeclareMathOperator*{\argmin}{arg\,min}

\usepackage{float}
\usepackage[caption = false]{subfig}
\usepackage[numbers,sort&compress]{natbib}
\setcitestyle{square, comma, numbers,sort&compress, super}
\usepackage{multirow}
\usepackage{booktabs,siunitx}
\usepackage{adjustbox}
\usepackage{stfloats}

\makeatletter
\AtBeginDocument{\DeclareMathVersion{bold}
\SetSymbolFont{operators}{bold}{T1}{times}{b}{n}
\SetSymbolFont{NewLetters}{bold}{T1}{times}{b}{it}
\SetMathAlphabet{\mathrm}{bold}{T1}{times}{b}{n}
\SetMathAlphabet{\mathit}{bold}{T1}{times}{b}{it}
\SetMathAlphabet{\mathbf}{bold}{T1}{times}{b}{n}
\SetMathAlphabet{\mathtt}{bold}{OT1}{pcr}{b}{n}
\SetSymbolFont{symbols}{bold}{OMS}{cmsy}{b}{n}
\renewcommand\boldmath{\@nomath\boldmath\mathversion{bold}}}
\makeatother

\def\BibTeX{{\rm B\kern-.05em{\sc i\kern-.025em b}\kern-.08em
    T\kern-.1667em\lower.7ex\hbox{E}\kern-.125emX}}

\begin{document}
\history{Date of publication xxxx 00, 0000, date of current version xxxx 00, 0000.}
\doi{10.1109/ACCESS.2023.1120000}

\title{GReAT: A Graph Regularized Adversarial Training Method}
\author{\uppercase{Samet Bayram}\authorrefmark{1}, \IEEEmembership{Member, IEEE},
\uppercase{Kenneth Barner}\authorrefmark{2},
\IEEEmembership{Fellow, IEEE}}

\address[1]{Electrical and Computer Engineering Department, University of Delaware, Newark, DE 19716 USA (e-mail: sbayram@udel.edu)}
\address[2]{Electrical and Computer Engineering Department, University of Delaware, Newark, DE 19716 USA (e-mail: barner@udel.edu)}
%\tfootnote{This paragraph of the first footnote will contain support
%information, including sponsor and financial support acknowledgment. For
%example, ``This work was supported in part by the U.S. Department of
%Commerce under Grant BS123456.''}

\markboth
{Bayram \headeretal: GReAT: A Graph Regularized Adversarial Training Method}
{Bayram \headeretal: GReAT: A Graph Regularized Adversarial Training Method}

\corresp{Corresponding author: Samet Bayram (e-mail: sbayram@udel.edu).}

\begin{abstract}
This paper presents GReAT (Graph Regularized Adversarial Training), a novel regularization method designed to enhance the robust classification performance of deep learning models. Adversarial examples, characterized by subtle perturbations that can mislead models, pose a significant challenge in machine learning. Although adversarial training is effective in defending against such attacks, it often overlooks the underlying data structure. In response, GReAT integrates graph-based regularization into the adversarial training process, leveraging the data's inherent structure to enhance model robustness. By incorporating graph information during training, GReAT defends against adversarial attacks and improves generalization to unseen data. Extensive evaluations on benchmark datasets demonstrate that GReAT outperforms state-of-the-art methods in robustness, achieving notable improvements in classification accuracy. Specifically, compared to the second-best methods, GReAT achieves a performance increase of approximately 4.87\% for CIFAR-10 against FGSM attack and 10.57\% for SVHN against FGSM attack. Additionally, for CIFAR-10, GReAT demonstrates a performance increase of approximately 11.05\% against PGD attack, and for SVHN, a 5.54\% increase against PGD attack. This paper provides detailed insights into the proposed methodology, including numerical results and comparisons with existing approaches, highlighting the significant impact of GReAT in advancing the performance of deep learning models.

%Extensive evaluations on benchmark datasets demonstrate that GReAT outperforms state-of-the-art methods in robustness, achieving notable improvements in classification accuracy. This paper provides detailed insights into the proposed methodology, including numerical results and comparisons with existing approaches, highlighting the significant impact of GReAT in advancing the performance of deep learning models. %
\end{abstract}


\titlepgskip=-21pt  %-21pt

\maketitle

\section{Introduction}
\label{sec:introduction}
\PARstart{D}{eep} learning is a subset of machine learning (ML) that uses artificial neural networks with multiple layers and neurons to analyze and learn from large amounts of data. Deep learning algorithms automatically learn and extract relevant features from the data to make predictions. The feature extraction process allows algorithms to achieve higher levels of accuracy and perform more complex tasks. In recent decades, deep learning has achieved impressive results in various domains, including image and text classification, speech recognition, image generation, and natural language processing~\cite{krizhevsky2012imagenet}, \cite{ liang2015recurrent}, \cite{goodfellow2020generative}. Supervised learning methods achieved the most successful results. In this learning technique, the model is trained on a labeled dataset, meaning the input data is accompanied by its corresponding output labels. Supervised learning aims to predict new, unseen data based on the patterns learned from the labeled data. Deep neural networks adjust the weights and biases of the network through back-propagation using output labels and original labels.

Semi-supervised learning combines both supervised and unsupervised learning techniques. In this type of learning, a model is trained on a data set that has labeled and unlabeled instances. The goal is to use the labeled data to make predictions on unlabeled and new, unseen, data. This method is used often when there is a limited amount of labeled data but a larger amount of unlabeled data. There are various algorithms for propagating labels through the graph, such as label propagation~\cite{bengio_label_2006, yang_revisiting_2016}, pseudo-labeling~\cite{lee_pseudo-label_2013}, transductive SVMs~\cite{joachims99transductive}, and self-training~\cite{selftraining}. Label propagation provides outstanding performance for semi-supervised learning to classify graph nodes. It is based on the idea that a node's labels can be propagated to its neighbors based on the assumption that nodes with similar labels are more likely to be connected.

Despite their significant success, deep learning models are known to be vulnerable to adversarial examples. These examples are created by adding small, carefully chosen perturbations to the input data. The perturbed data remain visually similar to the original data, but are misclassified by the model~\cite{harnesing, carlini17, nguyen2015deep}. The existence of adversarial examples has drawn significant attention to the machine-learning community. Showing the vulnerabilities of machine learning algorithms has opened critical research areas in the attack and robustness areas. Studies have shown that adversarial attacks are highly effective on many existing AI systems, especially on image classification tasks;~\cite{szegedy2014intriguing,Biggio_2018,carlini17,moosavidezfooli2016deepfool,sharif,kurakin16,eykholt2018robust,ecoba}. For instance, 
%Szegedy \emph{et al.}~
\cite{szegedy2014intriguing} show that even small perturbations in input testing images can significantly change the classification accuracy.

%$Goodfellow \emph{et al.}~
The authors of \cite{harnesing} attempt to explain the existence of adversarial examples and proposes one of the first efficient attack algorithms in white--box settings. 
%Madry \emph{et al.}~
\cite{madry2019deep} proposed projected gradient descent (PGD) as a universal first-order adversarial attack. They stated that network architecture and capacity play a significant role in adversarial robustness. Notable other popular adversarial attack methods are the Carlini-Wagner Attack (CW)~\cite{carlini17}, Basic Iterative Method (BIM)~\cite{kurakin16}, and Momentum Iterative Attack~\cite{papernot2016limitations}. 
%Tramèr \emph{et al.}~
\cite{transfer} show the transferability of black-box attacks among different ML models. 

%Adversarial defense algorithms can be considered into two categories. The first category focuses mainly on the pre-processing steps of DL models that aim to remove adversarial perturbations from the adversarial examples, such as feature denoising~\cite{xie2019feature}, Fourier filtering~\cite{bafna2018thwarting}, random resizing and random padding~\cite{xie2017mitigating}, \emph{etc.} The second category targets the Neural network structure such as activation functions~\cite{wang2018adversarial}, learning processes such as distillation~\cite{papernot2016distillation}, designing new loss function~\cite{chen2019improving}, or or manipulating training procedure~\cite{harnesing}.

Adversarial defense mechanisms can broadly be classified into three categories. The first category predominantly centers on pre-processing techniques tailored for DL models to mitigate adversarial perturbations in the adversarial examples. Methods in this category include feature denoising~\cite{xie2019feature}, Fourier filtering~\cite{bafna2018thwarting} and random resizing coupled with random padding~\cite{xie2017mitigating} among others. The second category targets modifications in the architecture of neural networks, including alterations in activation functions~\cite{wang2018adversarial}, adaptations to learning processes such as distillation~\cite{papernot2016distillation}, the introduction of novel loss functions~\cite{chen2019improving} and adjustments in training procedures~\cite{harnesing}. The last category attempts to implement the combination of the first two categories. 

The authors of~\cite{benchmark_sota} introduce two adversarial training methods, topology aligning adversarial training (TAAT) and adversarial topology aligning (ATA), which leverage topology information to maintain consistency in the topological structure within the feature space of both natural and adversarial examples, and further introduce a Knowledge-Guided (KG) training scheme to ensure stability and efficiency in topology alignment. TRADES (tradeoff-inspired adversarial defense via surrogate-loss minimization) ~\cite{TRADES} focuses on decomposing the prediction error for adversarial examples into natural error and boundary error. The authors also introduce a differentiable upper bound using the theory of classification-calibrated loss, which serves as the foundation for their defense method, TRADES, designed to trade off adversarial robustness against accuracy. 

The authors of adversarial logit pairing (ALP)~\cite{ALP} introduce a technique called logit pairing, which encourages similarity between logits for pairs of examples, resulting in improved accuracy on adversarial examples compared to standard adversarial training~\cite{madry2019deep}. Triplet Loss Adversarial
(TLA) training~\cite{TLA2019} introduces a metric learning approach to regularize the representation space under attack, resulting in increased robustness of classifiers against adversarial attacks. Adversarial Contrastive Learning (ACL)~\cite{ACL2020} enhances robustness-aware self-supervised pre-training by incorporating adversarial perturbations into a contrastive learning framework. This approach improves feature consistency under both data augmentations and adversarial perturbations, leading to models that are label-efficient and robust.

The existing literature underscores the effectiveness of unlabeled samples in improving deep learning performance~\cite{zhou_learning_2005,zhu2005phdthesis,belkin_manifold_2006,bengio_label_2006}. Additionally, studies show that unlabeled data improve adversarial robustness,~\cite{carmon_unlabeled_2019}. Motivated by these insights, we propose a Graph-Regularized Adversarial Training method (GReAT) to improve the robustness performance. The proposed method utilizes the structural information from the input data to improve the robustness of deep learning models against adversarial attacks. The main idea within GReAT is to construct a graph representation of clean data with an adversarial neighborhood, where each node represents a data point, and the edges encode the similarity between the nodes. This approach allows us to incorporate the structural information from the data into the training process, which helps create robust classification models. To evaluate the effectiveness of our approach, we conduct experiments on data sets: TensorFlow's flower data set~\cite{tfflowers} and CIFAR-10~\cite{cifar10}. We compare GReAT with several state-of-the-art methods. The results show that the proposed approach consistently outperforms the baselines in terms of accuracy and robustness against adversarial attacks. Our proposed GReAT graph-based semi-supervised learning approach for adversarial training provides a promising direction for improving the robustness of deep learning models against adversarial attacks. We list the main contributions of our proposed method below:
\begin{itemize}
    \item  GReAT integrates graph structure into the adversarial training process.
    \item  It improves the model's feature extraction performance by including neighboring information. 
    \item  The model enhances the learning capabilities of the model by leveraging the underlying structure of the training samples.
    \item The proposed method shows advantages in the adversarial training method, compared with the state-of-the-art methods, thereby improving generalization and robustness.

\end{itemize}

\section{Background and Related Works}
This section covers the relevant background and related works. In particular, we cover deep learning and semi-supervised learning, adversarial learning, and graph-based semi-supervised learning.

\subsection{Deep Learning and Semi-supervised Learning}

DL models are complex non-linear mapping functions between input and output. They consist of multiple layers and neurons with activation functions. They extract features from input samples and predict labels based on those features. Neural networks are trained using vast amounts of labeled data and can learn and improve their performance over time utilizing backpropagation algorithms. 

The following equation represents the prediction process of the classical deep learning paradigm:
\begin{equation}\label{eq:supervised}
    \boldsymbol{Y}: f(\boldsymbol{X},\boldsymbol{\theta},\boldsymbol{b}),
\end{equation}
where $\boldsymbol{X}$ is the data fed into the neural network $f$, $\boldsymbol{\theta}$ represent the values assigned to the connections between the neurons in the network, and  $\boldsymbol{b}$ is the offsets applied to the input data. The output is the result produced by the neural network after processing the input data through its layers of neurons. 

Semi-supervised learning uses labeled and unlabeled data to train a model. The following representation shows the prediction process of semi-supervised learning:
\begin{equation}\label{eq:ssl}
    \boldsymbol{Y}: f(\boldsymbol{X_l}, \boldsymbol{X_{ul}},\boldsymbol{\theta},\boldsymbol{b}),
\end{equation}
where $\boldsymbol{X_l}$ is the labeled data and $\boldsymbol{X_{ul}}$ is the unlabeled data. The weights and biases are the same as in supervised-learning learning. The output is the result produced by the model after processing the labeled and unlabeled data through its layers of neurons. A label assignment procedure typically exists in semi-supervised learning to annotate the unlabeled data. This procedure employs a smoothing function or similarity metrics to assign the label of the most similar labeled sample to the unlabeled sample,~\cite{yang_revisiting_2016}.

\subsection{Adversarial Learning}

A data instance $\boldsymbol{x}'$ is considered an adversarial example of a natural instance $\boldsymbol{x}$ when $\boldsymbol{x}'$ is close to $\boldsymbol{x}$, under a specific distance metric, while $f(\boldsymbol{x}') \neq y$, where $y$ is the label of $\boldsymbol{x}$, \cite{ren2020adversarial}. Formally, an adversarial example of $\boldsymbol{x}$ is can be defined as 
\begin{equation}\label{eq:adv_ex}
    \boldsymbol{x}': D(\boldsymbol{x}',\boldsymbol{x}) < \epsilon, f(\boldsymbol{x}') \neq y,
\end{equation}
where $D(\cdot,\cdot)$ represents a distance metric, such as the $\lVert \cdot \rVert_2$ norm, and $\epsilon$ is a distance constraint, which limits the amount of allowed perturbations.
Since the existence of adversarial examples is a significant threat to DL models, adversarial attack and defense algorithms are intensively investigated to improve the robustness and security of such models.

For instance, FGSM ~\cite{harnesing} was proposed to generate adversarial examples and attack DL models. The PGD algorithm, an iterative version of the FGSM attack, was proposed to generate adversarial examples by maximizing the loss increment within an $L_{\infty}$ norm-ball,~\cite{madry2019deep}.  Although many defense methods have been proposed, adversarial training is the most efficient approach against adversarial attacks,~\cite{harnesing, madry2019deep}. The authors of \cite{harnesing} proposed using adversarial attack samples during training so that the classifier can learn the features of adversarial examples and their perturbations. The classifier's robustness against adversarial attacks is substantially enhanced due to the integration of adversarial examples in the training phase. It effectively empowers the classifier to develop a more robust defense mechanism against adversarial instances. Formally, adversarial training is defined as
\begin{equation}\label{eq:adv_tr}
    \boldsymbol{\theta^*} = \argmin_{\theta \in \Theta}\frac{1}{L}\sum_{i=1}^L\max_{\textsf{D}(\boldsymbol{x}'_i,\boldsymbol{x}_i)<\epsilon}\ell_{adv}(\theta,\boldsymbol{x}'_i,y_i).
\end{equation}

The above equation states a min-max procedure under the specific distance constraint. In the inner maximization component, the adversarial training seeks an adversarial example $\boldsymbol{x}'_j$ to maximize the loss $\ell_{adv}$, under the distance metric $D(\boldsymbol{x}'_j,\boldsymbol{x}_j)<\epsilon$, given the natural sample $\boldsymbol{x}_j$. The outer minimization seeks the optimal gradient $\theta^*$ that yields the global minimum empirical loss. In their work, Madry~\emph{et al.} iteratively applies the PGD algorithm during training to search for strong adversarial examples to maximize $\ell_{adv}$. This helps the model yield improved robustness against PGD and FGSM attacks. Adversarial training with PGD is considered one of the strongest defense methods,~\cite{madry2019deep}.

\subsection{Graph-based Semi-supervised Learning}

Graph-based semi-supervised learning uses labeled and unlabeled data to train DL models,~\cite{zhou_learning_2005, belkin_manifold_2006, weston_deep_2012, agarwal_social_2009, jacob_learning_2014, bui_neural_2018}. This approach uses a small amount of labeled data and a large amount of unlabeled data to learn the graph structure of the given data. A given graph can be represented as $G = (V, E, W)$, where $V$ indicates data points as vertices, $E$ represents edges between data points, and $W$ is the edge weight matrix. The edges between the vertices are created on the basis of a similarity metric between the data points. Graph-based semi-supervised learning aims to use the graph structure and the labeled data to learn the label for the unlabeled data points. This technique is typically done by propagating the labels from the labeled data points to the unlabeled data points through the similarity graph of the entire data,~\cite{zhu2005phdthesis, zhou_learning_2004, yang_revisiting_2016, bui_neural_2018}. 

In graph-based semi-supervised learning, label propagation is often used to classify nodes in a graph when only a few nodes have been labeled. This method starts with the labeled nodes and propagates their labels to their neighbors. The labels are then iteratively propagated repeatedly until the mapping function converges and the entire graph is labeled. As shown in~\cite{yang_revisiting_2016}, the loss function of the graph-based semi-supervised learning can be represented as:
\begin{equation}
\label{eq:graphSSL}
     \sum_{i=1}^L\ell(\theta,\boldsymbol{x}_i,y_i)+ \lambda\sum_{i,j} w_{i,j}\lVert h(x_i)-h(x_j) \rVert^2,
\end{equation}
where the first term represents the standard supervised loss while the second term represents the penalty of the neighborhood loss. Note that $w_{ij}$ represents the similarity between different instances, and $\lambda$ controls the contribution of neighborhood regularization. When $\lambda=0$, the loss term becomes the standard supervised loss. The amount of penalty depends on the similarity between the instance $\boldsymbol{x_i}$ and its neighbors. In addition, $h$ represents a lookup table that contains all samples and similarity weights. It can be obtained with a closed-form solution, according to~\cite{zhou_learning_2004}. 

The authors of \cite{weston_deep_2012} proposed embedding samples instead of using lookup tables by extending the regularization term in Eq.~\ref{eq:graphSSL}. The regularization term becomes $ \lambda\sum_{i,j}\alpha_{i,j}\lVert g_{\theta}(x_i)-g_{\theta}(x_j) \rVert^2$, where $g_{\theta}(\cdot)$ indicates the embedding of samples generated by a neural network. Transforming the regularization term by transitioning from $f$ to $g$ leverages stronger constraints on the neural network, according to~\cite{yang_revisiting_2016}. 

\begin{figure*}[ht]
\begin{minipage}{1\linewidth}
  \centering
  \includegraphics[width=0.8\textwidth]{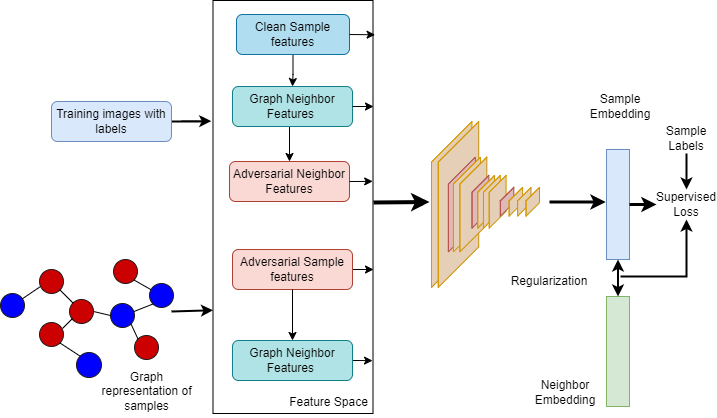}
%  \vspace{2.0cm}
  \centerline{}%\medskip
\end{minipage}
\caption{GReAT framework.}
\label{fig:great_framework}
\end{figure*}

Here, we extend Eq.~\ref{eq:graphSSL} by replacing the lookup table term with the embedding term in the regularization component and defining a general neighbor similarity metric. This approach yields Eq.~\ref{eq:graphSSL2} as 
\begin{equation}
\label{eq:graphSSL2}
     \sum_{i=1}^L\ell(\theta,\boldsymbol{x}_i,y_i)+ \lambda\sum_{i=1}w_i \textsf{D}(g_{\theta}(x_i), \textsf{N} (g_{\theta}(x_i))).
\end{equation}
In the Eq. \ref{eq:graphSSL2}, \textsf{N} represents the neighbors of a given sample $\boldsymbol{x_i}$, $\boldsymbol{w_i}$ represents the edge weight between the sample $\boldsymbol{x_i}$ and its neighbors, and D represents the distance metric between embeddings. 

\section{Graph Regularized Adversarial Training}
\label{sec:proposed methods}
In this Section, we integrate the adversarial learning process,~\cite{madry2019deep}, into the graph-based semi-supervised learning framework,~\cite{weston_deep_2012, yang_revisiting_2016, bui_neural_2018}, to take advantage of both adversarial training and semi-supervised learning. The main framework of GReAT is shown in Fig.~\ref{fig:great_framework}.
%%% figure 1 place

The feature space encompasses both the labeled original training samples and the adversarial examples that are created through adversarial regularization and neighbor similarities. This feature space is crucial for identifying the nearest-neighbor samples. When we feed a batch of input samples to the neural network, it includes not only the original samples but also their corresponding neighbors. In the final layer of the neural network, we derive a sample embedding for each of these samples. The training objective for regularization includes two components: the supervised loss and the label propagation loss, which accounts for neighbor-related loss. In other words, it considers the impact of neighbors on the overall training objective. Thus,
\begin{equation}
\label{eq:great_simple}
     \mathcal{L}_{GReAT} =\mathcal{L}_{adv} + \lambda \mathcal{L}_{N} ,
\end{equation}
where $\mathcal{L}_{adv}$ represents the supervised loss from training labels of clean and adversarially perturbed samples, and $\mathcal{L}_{N}$ represents the neighbor loss, which includes the loss from the clean training samples and adversarially perturbed samples. 

We consider similar instances as neighbors of sample $\boldsymbol{x}$ in the graph regularized semi-supervised learning case. In our case, we consider an adversarial example, $\boldsymbol{x'}$, in addition to a neighbor of sample $\boldsymbol{x}$. Next, we extend Eq.~\ref{eq:graphSSL2} by including adversarial and adversarial neighbor losses as new regularizer terms. Formally, the unpacked form of Eq.~\ref{eq:great_simple} is:
\begin{equation}\label{eq:great}    
\begin{split}
\mathcal{L}_{GReAT} (\Theta) = &\sum_{i=1}^L\ell(\theta,\boldsymbol{x}_i,y_i)\\
    &+\alpha_{11}\sum_{i=1}^L\ell_N(y_i, x_i, \textsf{N}(x_i))\\
    &+\alpha_{22}\sum_{i=1}^L\ell_N(y_i, x'_i, \textsf{N}(x'_i))\\
    &+\alpha_3\sum_{i=1}^L\ell(\theta, \textsf{N}_{adv}(x_i),y_i).
 \end{split}
\end{equation}

In the above equation, $\textsf{N}(\boldsymbol{x})$ represents neighbors of sample $\boldsymbol{x}$. The neighbors could be clean or adversarially perturbed samples. Thus, $\textsf{N}(\boldsymbol{x'})$ represents the neighbors of adversarial example $\boldsymbol{x'}$. Its neighbors could be clean samples and adversarial examples. Specifically, $\textsf{N}_{adv}(\boldsymbol{x})$ represents the adversarial neighbor of the sample $\boldsymbol{x}$. The adversarial neighbors have the same label as the original sample $\boldsymbol{x}$ similar to the standard adversarial training. 

\begin{figure*}[t]
\begin{minipage}[t]{1.0\linewidth}
  \centering
  \centerline{\includegraphics[width=0.85\textwidth]{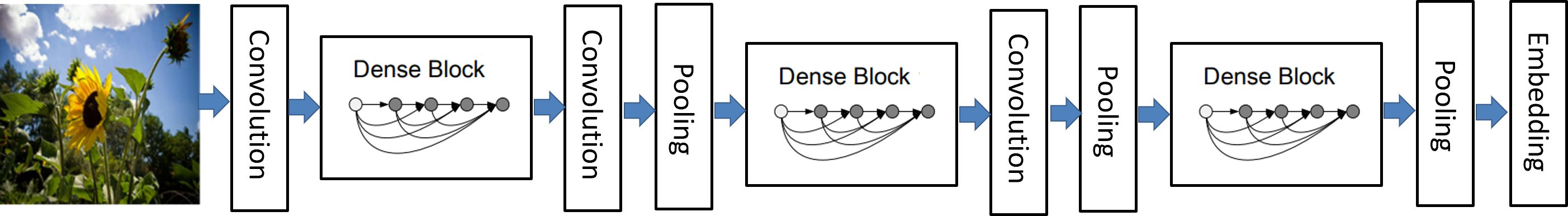}}
 % \vspace{2.0cm}
  \centerline{}\medskip
\end{minipage}
\caption{Densenet121 for generating image embeddings. }
\label{fig:graph_embeddings}
\end{figure*}

We obtain adversarial examples using  PGD as it's described in ~\cite{madry2019deep}. Note that the $\alpha_{11},\alpha_{22},\alpha_3$ hyperparameters determine the contributions of different neighborhood types, which are shown in Fig.~\ref{fig:tfflowers_subgraph} as sub-graph types. The $\alpha$ terms can be tuned according to the performance on clean and adversarially perturbed testing inputs. The pseudo-code of GReAT method is given in the Algorithm \ref{alg:great}. Furthermore, a detailed explanation of the embedding of neighbor nodes and graph construction between clean and adversarial examples is shown in Section~\ref{ssec:graph_construction}.

\begin{figure*}[b]
\begin{minipage}[b]{1.0\linewidth}
  \centering
  \centerline{\includegraphics[width=0.8\textwidth]{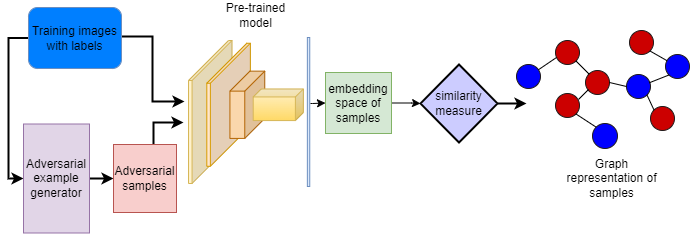}}
%  \vspace{2.0cm}
  \centerline{}\medskip
\end{minipage}
\caption{Graph creation from embedding of clean and adversarial examples. }
\label{fig:graph_embeddings_main}
\end{figure*}
%\subsection{Framework Overview}

\begin{algorithm}[H]
\caption{Graph Regularized Adversarial Training (GReAT)}
\label{alg:great}
\begin{algorithmic}[1]
\State \textbf{Input:} Labeled data $\boldsymbol{X_l}$, unlabeled data $\boldsymbol{X_{ul}}$, model parameters $\Theta$, hyperparameters $\alpha_{11}$, $\alpha_{22}$, $\alpha_3$, $\lambda$
\State \textbf{Output:} Trained model $\Theta^*$
\Statex
\State 1: Train classifier $f(\boldsymbol{X_l}, \Theta)$ on labeled data using supervised loss $\ell$
\State 2: Generate adversarial examples $\boldsymbol{X_{adv}}$ 
\State 3: Propagate labels of $\boldsymbol{X_l}$ to $\boldsymbol{X_{ul}}$ using label propagation on graph
\State 4: Construct graph $\mathcal{G}$ with nodes $\boldsymbol{X_l}, \boldsymbol{X_{ul}}, \boldsymbol{X_{adv}}$
\State 5: Compute neighbor set $\textsf{N}(\boldsymbol{x})$, $\textsf{N}(\boldsymbol{x'})$, $\textsf{N}_{adv}(\boldsymbol{x})$ for each sample $\boldsymbol{x}$, $\boldsymbol{x'}$
\State 6: Train model using loss $\mathcal{L}_{GReAT}$ defined in Eq.~\ref{eq:great}
\State \textbf{return} $\Theta^*$
\end{algorithmic}
\end{algorithm}

%%% figure 2 was here
%%

\subsection{Related Previous Methods}
\label{sec:format}
Creating graph embeddings using Deep Neural Networks (DNNs) is a well-known method,~\cite{weston_deep_2012}. Furthermore, the propagation of unlabeled graph embeddings using transductive methods,~\cite{zhu2005phdthesis, yang_revisiting_2016}, are efficient and well studied. Neural Graph Machines (NGMs),~\cite{bui_neural_2018}, are a commonly used example of label propagation and graph embeddings, along with supervised learning. The proposed training objective takes advantage of these frameworks and provides more robust image classifiers. Therefore, the training objective can be considered a combination of nonlinear label propagation and a graph-regularized version of adversarial training. 

\subsection{Graph Construction}
\label{ssec:graph_construction}
We use a pre-trained model, DenseNet121,~\cite{densenet121}, to generate image embeddings as a feature extractor. The pre-trained model has weights obtained by training on ImageNet. The pre-trained model is more complex than the model we use to train and test the proposed regularization algorithm in our simulations. Numerous studies show that complex DNNs are better feature extractors than shallow networks,~\cite{krizhevsky2009learning, shallow_nns}. Another significant advantage of using larger pre-trained models to obtain embeddings is to reduce computational costs. The process of creating embeddings is illustrated in Fig.~\ref{fig:graph_embeddings}.

\begin{figure*}[t]
\begin{minipage}{1.0\linewidth}
  \centering
  \centerline{\includegraphics[width=0.8\textwidth]{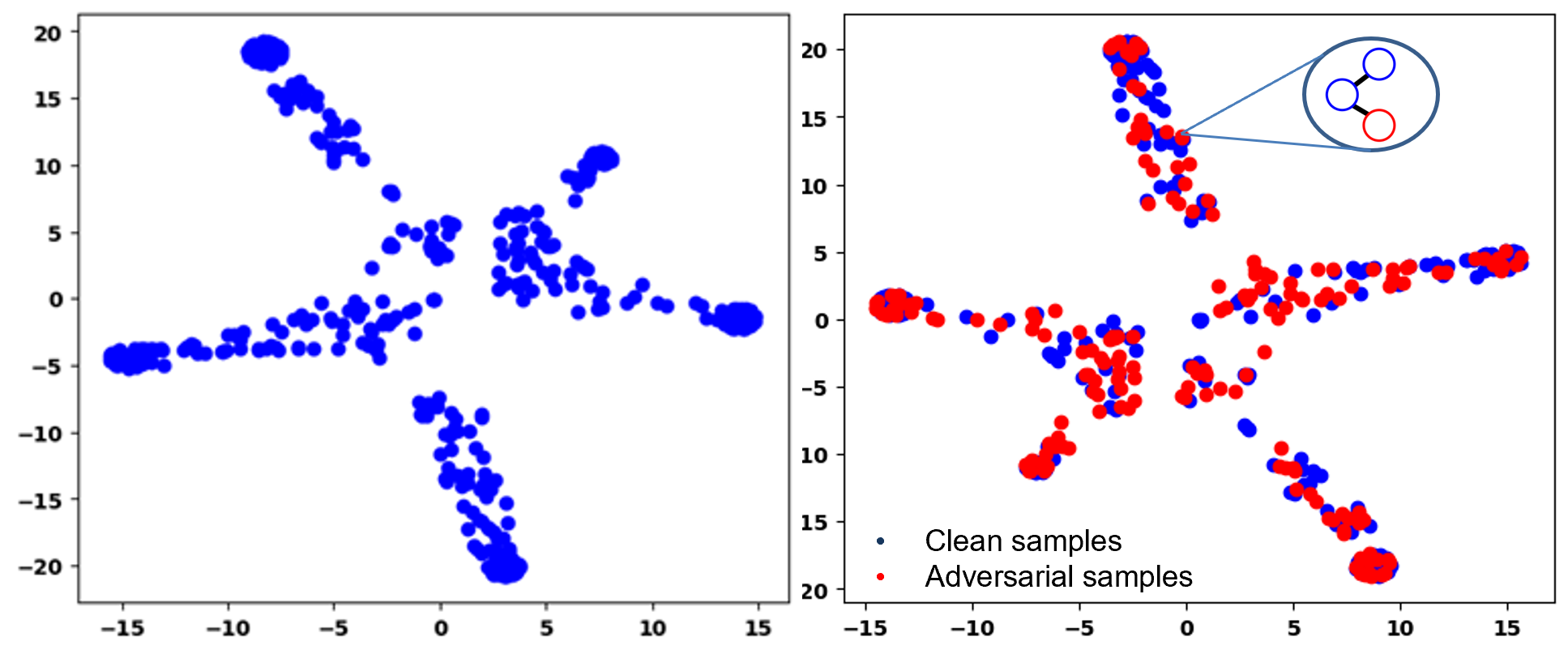}}
%  \vspace{2.0cm}
  \centerline{}\medskip
\end{minipage}
\caption{Samples in embedding space. The left figure represents all the samples in
the validation data set. The right figure shows some clean samples and their adversarial neighbors. }
\label{fig:t-sne}
\end{figure*}
%%%%%%%

\begin{figure*}[b]
\begin{minipage}[b]{1.0\linewidth}
  \centering
  \centerline{\includegraphics[width=0.65\textwidth]{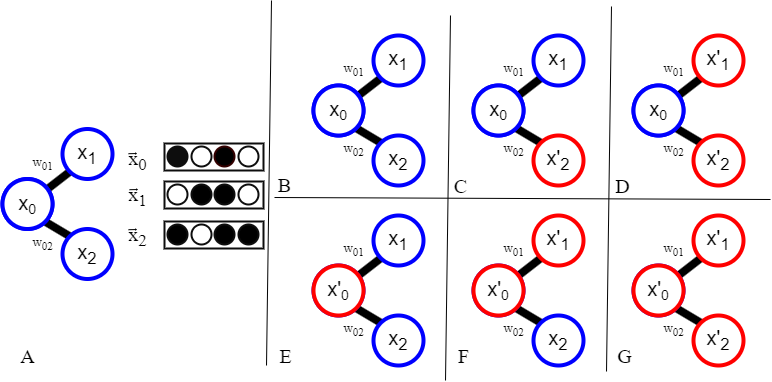}}
%  \vspace{2.0cm}
  %\centerline{}\medskip
\end{minipage}
\caption{A: A sample with two neighbors showing their sub-graph and feature inputs. Blue nodes represent clean samples, and red nodes represent adversarially perturbed samples. B,C,D,E,F, and G show how clean samples and adversarial examples may link on the graph structure.}
\label{fig:graphlets}
\end{figure*}

Generating appropriate inputs to the neural network plays a significant role in yielding correct predictions. As noted above, we use a pre-trained DL model to create node embeddings. We generate embeddings of clean samples and adversarial examples to obtain the neighborhood relationship between clean and adversarially perturbed examples. The overall graph construction process is shown in Fig.~\ref{fig:graph_embeddings_main}. Similarly, one-dimensional embedding is a crucial process for measuring sample similarities. Since the size of the embeddings is the same, we can visualize clean and adversarial examples in the embedding space using the \cite{tsne_maaten} t-distributed stochastic neighbor embedding (t-SNE) method.

% figure-4 was here

In Fig.~\ref{fig:t-sne}, we utilize t-SNE (t-Distributed Stochastic Neighbor Embedding) to create a visual representation of the validation data set obtained from TensorFlow's flower dataset. The primary purpose of this visualization is to provide insight into the distribution and relationships among the data points.The left panel of the figure is dedicated to displaying all the samples that constitute the validation data set. It is important to note that this data set encompasses samples belonging to five distinct classes. Each class represents a specific category or type of data within the dataset, and the samples within each class share certain common characteristics or features.

By visually representing the data set using t-SNE, we aim to reduce the dimensionality of the data while preserving its inherent structure and relationships. This reduction in dimensionality allows us to plot the data points in a two-dimensional space, making it easier to discern patterns, clusters, and similarities among the samples. Visualization is a valuable tool for gaining a deeper understanding of how the different classes are distributed and how they relate to each other within the validation data set.
The figure panel on the right shows how adversarial examples are distributed around clean samples. 

\begin{figure*}[t]
\begin{minipage}{1.0\linewidth}
  \centering
   \centerline{\includegraphics[width=0.8\textwidth]{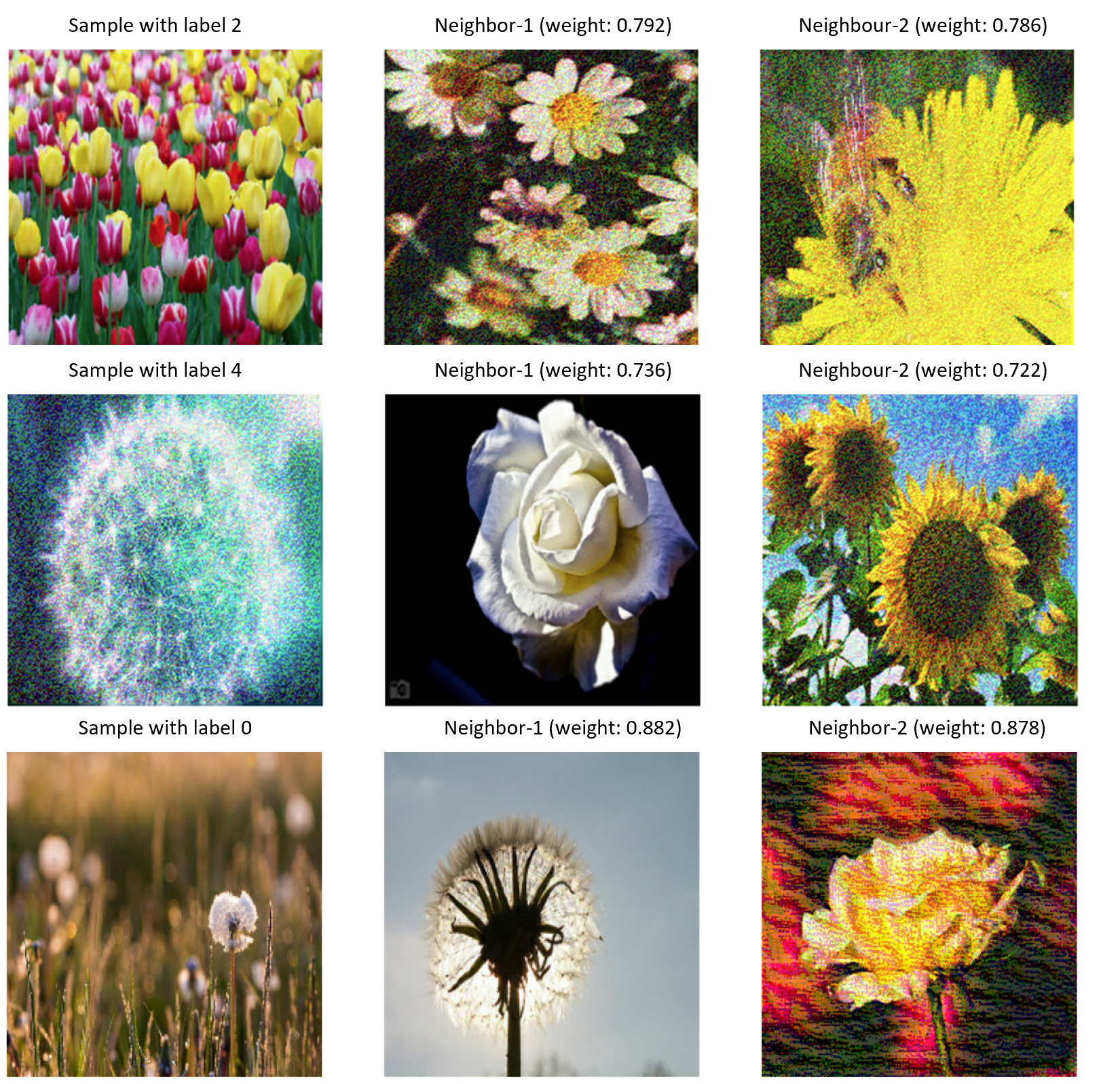}}
%  \vspace{2.0cm}
  %\centerline{}\medskip
\end{minipage}
\caption{From left to right: labeled sample, the first neighbour and the second neighbour. The samples are taken from Tensorflow's flowers data set.}
\label{fig:tfflowers_subgraph}
\end{figure*}

The visualization of the embeddings highlights a strong connection between individual samples and their respective neighbors, effectively distinguishing between various classes. We use the strong neighborhood connections to learn better and create more robust models. Consequently, we use these node embeddings as input features to the neural network by creating an adjacency embedding matrix, as shown in Fig.~\ref{fig:graphlets}. In particular, we use the label propagation method,~\cite{lee_pseudo-label_2013}, to propagate the information from the labeled data points to the unlabeled instances, which improves the model's performance on both clean and adversarial examples. 

Sample sub-graph of training instances are shown in Fig.~\ref{fig:graphlets}.  These examples might be labeled or unlabeled since we generate embeddings for each sample and create the graph based on the similarity between embeddings. A visual example of a sub-graph is demonstrated in Fig.~\ref{fig:tfflowers_subgraph}. Three examples of sub-graph types are shown. The first column of the figure shows labeled samples. The second and third columns show the labeled samples' two most similar neighbors. We associate these samples and their neighbors with the sub-graph examples, as noted in Fig.~\ref{fig:graphlets}. For instance, the first row of images in Fig.~\ref{fig:tfflowers_subgraph} represents Fig.~\ref{fig:graphlets}-D, since the labeled sample is clean and its first and second neighbors are adversarially perturbed samples. The second row of Fig.~\ref{fig:tfflowers_subgraph} represents Fig.~\ref{fig:graphlets}-F, since the labeled sample is adversarially perturbed and its neighbors are one clean sample and one adversarially perturbed sample. Finally, the third row represents Fig.~\ref{fig:graphlets}-C, since the labeled instance is clean and its neighbors are clean and adversarially perturbed samples. 

Note that a labeled sample may have one neighbor or none, for instance, if the similarity measure of the embeddings cannot pass the similarity threshold. In that case, the labeled sample goes through the neural network as a regular input without graph regularization.  

% FIGURE5 was here

% FIGURE 6 was here

%%% provide more sub-graph examples in the Apendix

\subsection{Optimization}
\label{ssec:opt}
%SGD, mini-batch, and the loss function .
The training process begins with a minibatch of samples and their edges. Instead of using all available data at once, the training process randomly selects a subset of edges for each iteration. This helps introduce randomness and variability into the training process, which is beneficial to the learning process. Additionally, to further improve the training process, selected edges are chosen from a nearby region to increase the likelihood of some edges. This can help reduce noise and speed up the learning process. As was implemented in other benchmark models \cite{benchmark_sota}, The Stochastic Gradient Descent (SGD) algorithm updates network weights utilizing the cross-entropy loss function.  

Note that the overall open form of the cost function in the following is equivalent to Eq.~\ref{eq:graphSSL2}. The cost function incorporates the cost of supervised loss from labeled clean and labeled adversarial examples and neighbor losses. That is, the cost includes different neighbor types/edges, as shown in Fig.~\ref{fig:graphlets}. Formally,
\begin{equation}\label{eq:great_detailed}    
\begin{split}
\mathcal{L}_{GReAT} (\Theta) = &\sum_{i=1}^L\ell(\theta,\boldsymbol{x}_i,y_i) +\sum_{i=1}^L\ell(\theta,\boldsymbol{x'}_i,y_i),\\
    &+\lambda\Biggr[\alpha_{11}\sum_{i=1}^Lw_{a} \textsf{D}(g_{\theta}(x_i), \textsf{N} (g_{\theta}(x_i)))\\
     &+\alpha_{12}\sum_{i=1}^Lw_{b} \textsf{D}(g_{\theta}(x_i), \textsf{N} (g_{\theta}(x'_i)))\\ 
     &+\alpha_{21}\sum_{i=1}^Lw_{c} \textsf{D}(g_{\theta}(x'_i), \textsf{N} (g_{\theta}(x_i)))\\ 
     &+\alpha_{22}\sum_{i=1}^Lw_{d} \textsf{D}(g_{\theta}(x'_i), \textsf{N} (g_{\theta}(x'_i)))\Biggr],
\end{split}
\end{equation}
where $w_a, w_b, w_c, w_d$ represent the similarity weights between the samples and their neighbors calculated by cosine similarity measurement. 

The similarity weights are (possibly) unique for each sample and its neighbors, with a range of zero to one. A sample and neighbor candidate are dissimilar if the similarity weight is near zero. For calculating the neighbor loss, we use $\textsf{D}$ as it represents the distance between a sample and its neighbor, where we use the norms $L_1$ and $L_2$ as distance metrics for calculating the neighbor distance. The hyperparameters $\alpha_{11}, \alpha_{12}, \alpha_{21}$ and $\alpha_{22}$ control the contributions of the different types of edges. For simulations, we set all $\alpha$s as one to include all edges in the training. The new objective function makes SGD possible with clean and adversarial examples and their neighbors in mini-batch training.

\subsection{Complexity Analysis}
\label{ssec:comp}
%%/The proposed method uses graph regularization on labeled and unlabeled data instances on the graph, which consists of benign and adversarial examples. The training complexity depends on the number of edges on the graph $E_c$. Therefore, we can show the complexity of each training epoch by $O(count(E_c))$. Note that $E_c$ is proportional to the number of neighboring data points considered. Furthermore, it also scales with the parameter that determines the selection of the most similar neighbors. Additionally, the size of the adversarial regularization step size is related to $E_c$.
%%/For instance, if we use only single-step adversarial regularization, FGSM, there will be one adversarial neighbor of the clear example. The multistep adversarial regularization, PGD, will significantly increase the number of edges, since we generate adversarial examples for each step. PGD-type adversarial regularization makes the model more robust than the FGSM regularization type. However, there is a trade-off between robustness and training time. Training the PGD regularized model is more resource intensive than FGSM models, since there are more edges and samples.

The proposed method incorporates graph regularization into its training process, applying it to both labeled and unlabeled data instances within the graph, which includes benign and adversarial examples. The computational complexity of each training epoch is dependent on the number of edges in the graph, denoted as $E_c$. To assess the complexity of the training, we can express it as $O(count(E_c))$. It is important to note that the quantity $E_c$ is directly proportional to several factors. First, it scales with the number of neighboring data points taken into account, indicating that more neighbors will increase the complexity. Second, it is influenced by a parameter that determines the selection of the most similar neighbors, further impacting the computational load. Furthermore, the step size used for adversarial regularization is tied to $E_c$.

For instance, if we opt for a single-step adversarial regularization method like FGSM, each clear example will have only one adversarial neighbor. However, when employing a multi-step adversarial regularization approach, the number of edges substantially increases, as adversarial examples are generated at each step. This type of PGD-based adversarial regularization tends to enhance the model's robustness compared to FGSM regularization. Nevertheless, it introduces a trade-off between robustness and training time. Training a model with PGD regularization demands more computational resources because of the increase in the number of edges and samples involved. This trade-off is essential when choosing the appropriate adversarial regularization method for a given application. For our simulations, we used FGSM to create adversarial examples for training and testing stages to reduce computational time.

\section{Experiments}
\label{sec:sims}
We conducted experiments to show the performance of the proposed GReAT method. Each experiment is carried out on clean data sets with a fixed number of epochs and training steps. The typical hyperparameters are fixed to ensure fair comparisons with other state-of-the-art methods. The base CNN model is trained and then regularized with the proposed loss function. We use the copy of the base model to obtain the regularized model each time to preserve the original base model. Once the models are trained, we test each model on the same clean and adversarially perturbed test data to measure the generalization and robustness performances.

\subsection{Datasets}
\label{ssec:dataset}

The Canadian Institute for Advanced Research dataset (Cifar10)~\cite{cifar10}, The Street View House Numbers (SVHN)~\cite{svhn}, and flowers~\cite{tfflowers} dataset are used to evaluate the methods. The Cifar10 dataset consists of 60,000 images with ten classes, and each class contains a fixed size of $32\times 32$ three-channel RGB images. To further assess the robust generalization capabilities of our proposed method, we conduct evaluations using the SVHN dataset. SVHN comprises 73,257 training samples and 26,032 testing samples. The flowers dataset contains 3,670 images with five classes, each containing high-resolution RGB images. The image sizes are not fixed in the flowers dataset. Resizing is, therefore, required as one of the pre-processing steps. 

The image distributions of each class are balanced for both data sets. We split the dataset 80\%-10\%-10\%, as train-validation-test data sets, respectively. In the simulations, we use the flowers dataset only for the ablation study since the state-of-the-art methods on the benchmark do not use flowers. In the ablation study, we reduce the training set to 20\%, and 50\%, to observe the model performances with fewer labeled samples.      

\subsection{Pre-processing Steps}
A few essential pre-processing steps are required to prepare the batches for training. After creating image embeddings, we measure the similarity between each embedding and create training batches based on this similarity metric.
\subsubsection{Similarity measure}
Identifying the closest neighbors for a given sample requires the measurement of similarity amongst the embeddings. Various metrics are available, including Euclidean distance, cosine similarity, and Structural Similarity Index Measure (SSIM). We have opted for cosine similarity due to its proven effectiveness in quantifying the similarity of image embeddings within a multidimensional space. Formally defined, the cosine similarity of two vectors can be expressed as follows:
\begin{equation}
\label{eq:cosine_sim}
     Cos(x_i, x_j) = \frac{x_i \cdot x_j }{\lVert x_i\rVert * \lVert x_j\rVert }.
\end{equation}

The similarity weights are between 0 and 1, depending on the angle between the two vectors. Two overlapping embeddings have weight 1 when the angle between the two embeddings is zero. Conversely, if two embedding vectors are orthogonal, they are dissimilar, and the similarity weight is zero. Once all similarity weights are calculated, the most similar neighbors are identified as candidates for regularization. We pre-define a similarity threshold to consider those neighbors. Embeddings that fall under the threshold are not considered as neighboring candidates on the graph.    

\subsubsection{Training batches}
Once the graph structure is created with clean samples and adversarial examples, we generate training batches that are fed into the neural network model. Each training batch consists of samples, their neighbors, and adversarial neighbors. The number of neighbors is predetermined, although other strategies can be utilized. In our simulations, we pick the number of neighbors as two.
\subsubsection{Adversarial Examples Generation}
In the training phase, we adhere to the conventional configuration of the $\ell_\infty$ threat model with a radius of 8/255. Adversarial examples are crafted using the PGD attack, iterating 10 steps with a size of 2/255 for all datasets.

\subsection{Network}
\label{ssec:models}

For our experiments, we utilize ResNet-18 \cite{resnet18} as the default baseline model architecture. We employ the SGD optimizer for training all models, with a momentum of 0.9 and weight decay set to 0.0005. The training batch size is set to 128. We use the same baseline model architecture and training parameters with other methods for a fair comparison. Additionally, we included the GReAT model trained with Adam optimizer to the benchmark. We noticed that using the Adam optimizer performs better than SGD during our experiments. The Adam optimizer with a 0.001 learning rate is utilized.  

\subsection{Results}
\label{ssec:sim_results}

\subsubsection{Ablation Study}
\label{ssec:results_tf_flowers}

As mentioned earlier, we use the flowers dataset for the ablation study. Because the dataset is relatively small, we deployed a smaller network architecture for the ablation study. This training model consists of 6 convolution layers and max-pooling layers. Dropout and batch normalization layers in the base model are deployed to minimize over-fitting. Subsequently, experiments are conducted on both clean and adversarially perturbed datasets to gauge model generalization and robustness. We set the perturbation magnitude to 0.2 in these experiments and employ the FGSM attack method. We compare the results with the standard AT model \cite{madry2019deep} and standard graph regularized model \cite{bui_neural_2018} to show the performance improvements with varying training set sizes. The performance of the proposed method, along with other methods on the clean testing dataset, is summarized in Table~\ref{tab:flowr_clean_accuracy}.

\begin{table}[h!]
\centering
\caption{Clean Accuracy Results for Flowers Data Set}
\label{tab:flowr_clean_accuracy}
\resizebox{\linewidth}{!}{
    \begin{tabular}{ccccccc}
    \toprule
    \multicolumn{7}{c}{\textbf{Model Accuracy}} \\
    \cmidrule{1-7}
    &\textbf{train set(\%)} & \textbf{Base} & \textbf{NSL} & \textbf{AT} & $\textbf{GReAT}_{adv}$ & \textbf{GREAT} \\
    \midrule
                & 20\% & 0.548 & \textbf{0.553} & 0.525 & 0.207 & 0.550 \\
                & 50\% & 0.564 & 0.608 & 0.575 & 0.245 & \textbf{0.659} \\
                & 80\% & 0.597 & 0.613 & 0.583 & 0.277 & \textbf{0.671} \\
    \bottomrule
    \end{tabular}
}
\end{table}

%As demonstrated in Table~\ref{tab:flowr_clean_accuracy}, the approach employing NSL~\cite{bui_neural_2018} unsurprisingly exhibits superior performance. This can be attributed to the use of only clean samples and their neighbors within the training batches. For completeness, we have incorporated $GReAT_{adv}$, which trains exclusively on adversarial examples and their associated neighbors. This serves to determine the effects of adversarial regularization. However, due to the lack of clean samples in its training batch, the model encounters difficulties when tested on a clean dataset. However, the proposed GReAT method still manages to achieve commendable performance.

Table~\ref{tab:flowr_clean_accuracy} indicates that the NSL approach,~\cite{bui_neural_2018}, yields significantly better performance. This is largely due to its training on clean samples and their respective neighbors. For a comprehensive evaluation, we introduced $GReAT_{adv}$, which trains only adversarial examples and their neighbors to assess the impact of adversarial regularization. Given its exclusive focus on adversarial examples during training, this model faces challenges when tested on clean datasets. However, the proposed GReAT method consistently yields positive results.

%\vspace{0.5cm}
\begin{table}[h!]
\centering
\caption{Robust Accuracy for Flowers Data Set}
\label{tab:flowers_robust_accuracy}
\resizebox{\linewidth}{!}{%
    \begin{tabular}{ccccccc}
    \toprule
    \multicolumn{1}{c}{\textbf{Attack Norm}} & \multicolumn{1}{c}{\textbf{train set(\%)}} & \multicolumn{5}{c}{\textbf{Model Accuracy}} \\
    \cmidrule{3-7}
    & & \textbf{Base} & \textbf{NSL} & \textbf{AT} & $\textbf{GReAT}_{adv}$ & \textbf{GReAT} \\
    \midrule
    L2 & 20\% & 0.011 & 0.011 & 0.450 & \textbf{0.836} & 0.605 \\
                  & 50\% & 0.014 & 0.024 & 0.496 & \textbf{0.854} & 0.647 \\
                  & 80\% & 0.016 & 0.063 & 0.526 & \textbf{0.891} & 0.668 \\
    \midrule
    Linf & 20\% & 0.001 & 0.000 & 0.727 & \textbf{0.924} & 0.883 \\
                    & 50\% & 0.002 & 0.001 & 0.753 & \textbf{0.942} & 0.892 \\
                    & 80\% & 0.005 & 0.005 & 0.819 & \textbf{0.968} & 0.931 \\
    \bottomrule
    \end{tabular}

}
\end{table}

%The NSL~\cite{bui_neural_2018} approach achieves the best performance, unsurprisingly, in Table~\ref{results_cln_flowers} since it has only clean samples and their neighbors in the training batches. We included $GReAT_{adv}$ for completeness, which trains only adversarial examples and their neighbors to see the effect of adversarial regularization. Since it does not have clean samples in its training batch, the testing performed on a clean testing data set suffers. The proposed GReAT method, yet still achieves reasonably good performance.   

Finally, the models were evaluated using adversarially perturbed test data from the flowers data set, and the results are shown in Table \ref{tab:flowers_robust_accuracy}. As the table shows, models not trained on adversarial examples, particularly the base model, exhibit diminished performance. Although the NSL model is trained only on clean samples, it still exhibits some robustness to adversarially perturbed test samples. The proposed GReAT model outperforms the other models and provides a balanced result for clean and adversarially perturbed testing data. $GReAT_{adv}$ gives the highest accuracy for perturbed test samples. This experiment shows how graph regularization with adversarial training is effective on both adversarially perturbed and clean testing samples.

\subsubsection{Accuracy vs attack strength}
\label{ssec:results_Cifar10_chanign epsilon}

\begin{figure}[ht]
\subfloat{\includegraphics[width = 0.5\textwidth]{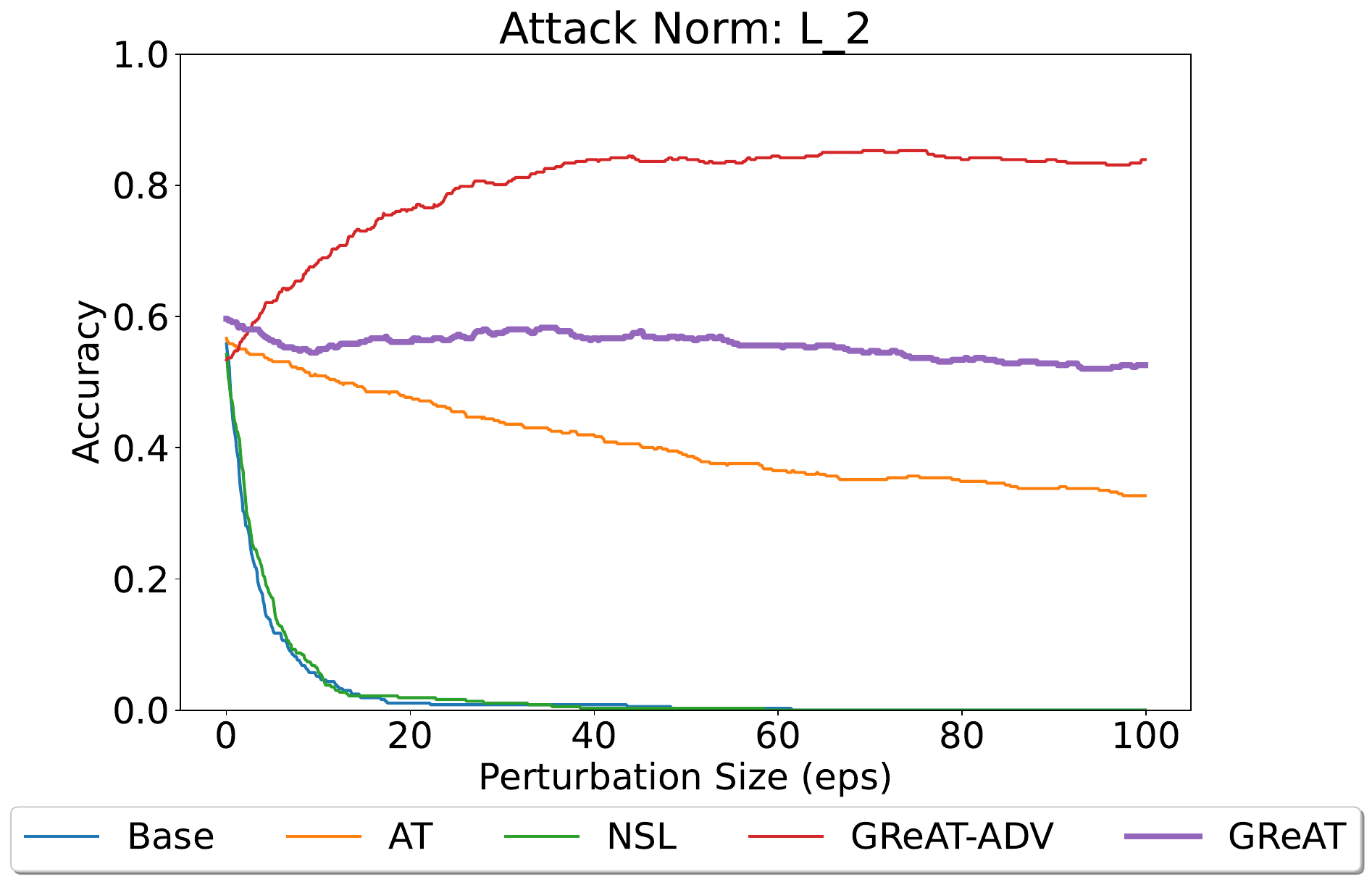}} \\
\subfloat{\includegraphics[width = 0.5\textwidth]{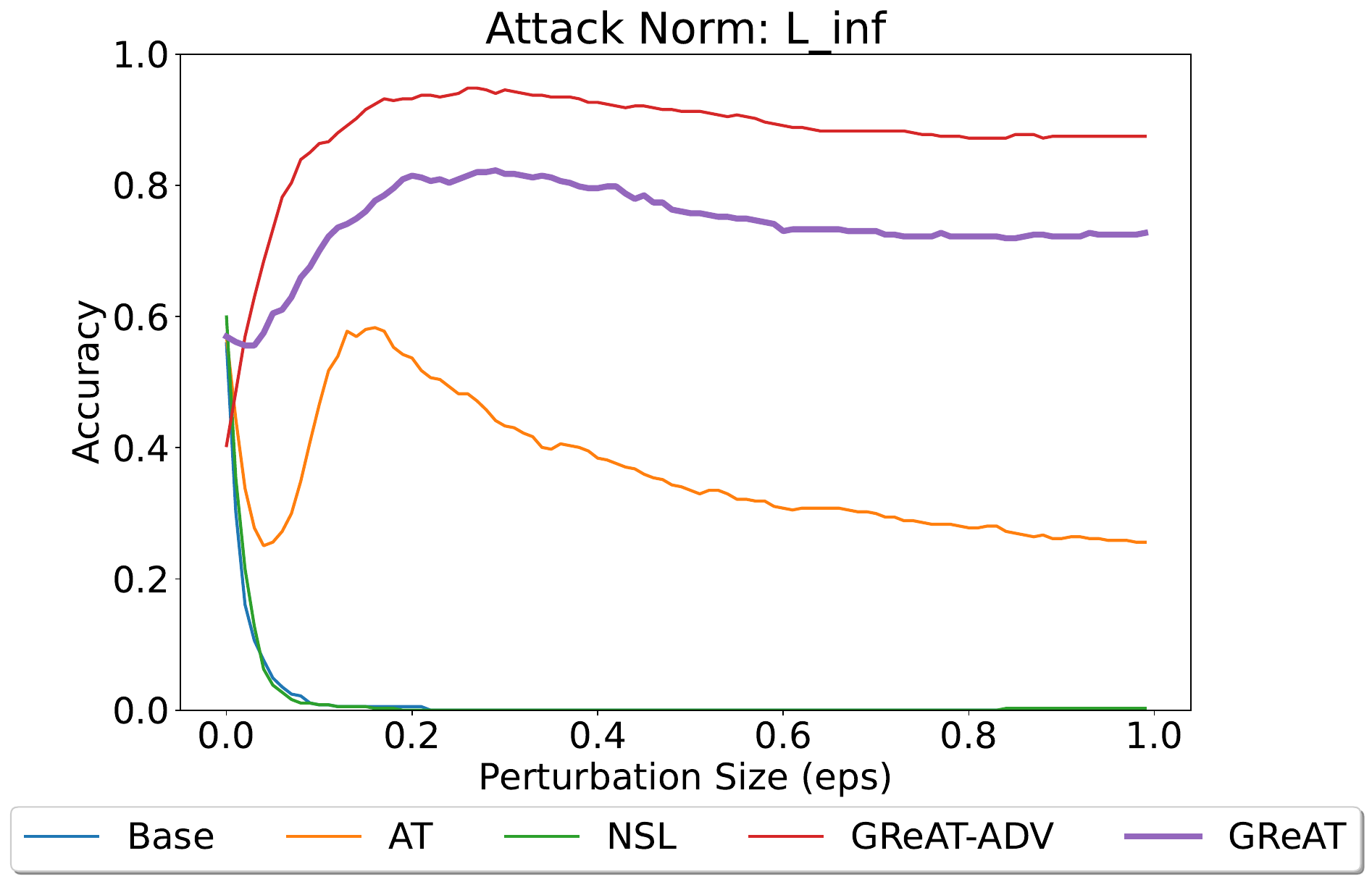}}

\caption{Robustness test with increasing perturbation size. }
\label{fig:eps_vs_perf}
\end{figure}

We evaluate the robustness of the proposed methods by adjusting the step size of the perturbations, which provides insights into the model performance under varying attack strengths. As illustrated in Fig.~\ref{fig:eps_vs_perf}, the accuracy of the base model declines sharply with increasing attack intensity. Although the model trained with standard adversarial training also exhibits a notable decrease in confidence, the proposed GReAT model consistently displays significant robustness, retaining its efficacy even under substantial perturbations.

The perturbation sizes for adversarial training samples are \(20\) for \(L_2\) norm and \(0.2\) for \(L_{\infty}\) constrained models, respectively. Training samples for \(L_2\) norm trained model are not normalized to one since we follow the training procedures as described in \cite{madry2019deep}. As the figure shows, the models trained on adversarial examples exhibit peak performance around these specific training perturbation sizes (\(\epsilon\)). This phenomenon occurs when the attack perturbation step size is smaller than the step size of the adversarial examples used for training \cite{madry2019deep}. Ideally, we aim to train a model with varying perturbation sizes to enhance its robustness against adversarial attacks, given that attack perturbation sizes might differ from the training perturbation size. However, for the sake of simulation simplicity, we utilize only a single-step perturbation size for the model training.

%%%%%%%%%%%%%%%%%%%% results for cifar10 dataset%%%%%%%%%%%%%%%%%%%%%%%%

\subsubsection{Ablation study on Cifar10 Data Set}
\label{ssec:results_Cifar10}
Next, we evaluate the performance on the Cifar10 dataset, which is made up of images with lower resolution with more classes and quantities. Tables~\ref{results_cln_cifar10} and \ref{tab:cifar10_robust_accuracy} detail the performance on the clean and adversarially perturbed image datasets, respectively. The proposed GReAT methodology yields balanced results, indicating that GReAT demonstrates both generalization and robust performance. In stark contrast, alternative methodologies are significantly impacted by adversarial attacks. These simulation results underscore that regularizing the deep learning model with both benign and adversarial examples results in improved generalization \emph{and} robustness.

\begin{table}[h!]
\centering
\caption{Clean Accuracy Results for Cifar10 Data Set.}
\label{results_cln_cifar10}
\resizebox{\linewidth}{!}{
    \begin{tabular}{ccccccc}
    \toprule
    \multicolumn{7}{c}{\textbf{Model Accuracy}} \\
    \cmidrule{1-7}
    &\textbf{train set(\%)} & \textbf{Base} & \textbf{NSL} & \textbf{AT} & $\textbf{GReAT}_{adv}$ & \textbf{GReAT} \\
    \midrule
     & 20\% & 0.522 & 0.523 & 0.296 & 0.227 & \textbf{0.560} \\
                & 50\% & 0.612 & 0.648 & 0.437 & 0.285 & \textbf{0.649} \\
                & 80\% & 0.701 & 0.713 & 0.688 & 0.327 & \textbf{0.731} \\
    \bottomrule
    \end{tabular}
}
\end{table}

Compared to other methods, the proposed method shows outstanding performance on the benign testing set. This is because more training data and classes provide more underlying information between classes with graph regularization. 

%\vspace{0.5cm}
\begin{table}[h!]
\centering
\caption{Robust Accuracy for Cifar10 Data Set.}
\label{tab:cifar10_robust_accuracy}
\resizebox{\linewidth}{!}{
    \begin{tabular}{ccccccc}
    \toprule
    \multicolumn{1}{c}{\textbf{Attack Norm}} & \multicolumn{1}{c}{\textbf{train set(\%)}} & \multicolumn{5}{c}{\textbf{Robust Accuracy}} \\
    \cmidrule{3-7}
    & & \textbf{Base} & \textbf{NSL} & \textbf{AT} & $\textbf{GReAT}_{adv}$ & \textbf{GReAT} \\
    \midrule
    L2            & 20\% & 0.121 & 0.161 & 0.343 & \textbf{0.556}& 0.525 \\
                  & 50\% & 0.090 & 0.172 & 0.367 & \textbf{0.594} & 0.567 \\
                  & 80\% & 0.135 & 0.191 & 0.385 & \textbf{0.651} & 0.638 \\
    \midrule
    Linf          & 20\% & 0.003 & 0.071 & 0.415 & \textbf{0.599} & 0.583 \\
                  & 50\% & 0.004 & 0.079 & 0.517 & \textbf{0.640} & 0.626 \\
                  & 80\% & 0.004 & 0.105 & 0.570 & \textbf{0.694} & 0.648 \\
    \bottomrule
    \end{tabular}

}
\end{table}

Table \ref{tab:cifar10_robust_accuracy} provides the performance of each model on adversarially perturbed testing data. As detailed in the table, the proposed method provides superior results to the NSL and standard adversarial training models. We observe similar results for $GReAT_{adv}$ in the Cifar10 data set, which shows the learning ability of GReAT over adversarial deceptive perturbations. 

\subsubsection{Comparison with the State-of-the-art Models}
Our proposed method's robust accuracy comparison results and various baseline models are presented under different attack methods, FGSM and PGD-100, on CIFAR-10 and SVHN datasets, using the $\ell_\infty$ norm with $\epsilon$ = 8/255. All models are based on the ResNet-18 architecture. The best checkpoint is selected based on the highest robust accuracy on the test set.

\begin{table}[h!]
\centering
\caption{Robust Benchmark under $l_{inf}$ type attack for CIFAR-10}
\label{results_sota_cifar10}
\resizebox{\linewidth}{!}{
    \begin{tabular}{ccccc}
    \toprule
    \multicolumn{4}{c}{\textbf{CIFAR-10, $l_{inf}=8/255$ , untargeted attack}} \\
    \cmidrule{1-4}
    &\textbf{Method} & \textbf{Natural Acc} & \textbf{FGSM} & \textbf{PGD-100}  \\
    \midrule
     & AT \cite{madry2019deep}    & 82.97 & 57.77 & 51.35  \\
     & ALP \cite{ALP}             & \textbf{84.86} & 57.55 & 51.57  \\
     & TLA \cite{TLA2019}         & 83.49 & 58.17 & 51.96  \\
     & ACL \cite{ACL2020}         & 83.26 & 57.54 & 51.51  \\
     & TRADES \cite{TRADES}       & 83.74 & 59.54 & 52.73  \\
     & ATA \cite{benchmark_sota}  & 83.41 & 57.96 & 52.39  \\
     & TAAT \cite{benchmark_sota} & 83.12 & 59.91 & 54.45  \\
     & GReAT (SGD)                      & 82.64 & \textbf{62.78} & \textbf{60.58}  \\
     & GReAT (ADAM)                     & 82.89 & \textbf{72.47} & \textbf{71.31}  \\
    \bottomrule
    \end{tabular}
}
\end{table}

The benchmark Table \ref{results_sota_cifar10} illustrates the robustness results for the CIFAR-10 dataset, showcasing the performance of various methods in defending against FGSM and PGD-100 adversarial attacks. Our proposed method notably outperforms existing approaches in terms of robust accuracy. However, it is observed that the natural accuracy achieved by our method is slightly lower compared to some other methods. Despite this, the significant improvement in robust accuracy demonstrates the effectiveness of our proposed method in enhancing the robustness of CIFAR-10 classification models against adversarial attacks.

\begin{table}[h!]
\centering
\caption{Robust Benchmark under $l_{inf}$ type attack for SVHN}
\label{results_sota_svhn}
\resizebox{\linewidth}{!}{
    \begin{tabular}{ccccc}
    \toprule
    \multicolumn{4}{c}{\textbf{SVHN, $l_{inf}=8/255$ , untargeted attack}} \\
    \cmidrule{1-4}
    &\textbf{Method} & \textbf{Natural Acc} & \textbf{FGSM} & \textbf{PGD-100}  \\
    \midrule
     & AT \cite{madry2019deep}    &  \textbf{90.5} & 65.08 & 52.87  \\
     & ALP \cite{ALP}             & 90.67 & 65.51 & 54.07  \\
     & TLA \cite{TLA2019}         & 90.63 & 64.66 & 52.96  \\
     & ACL \cite{ACL2020}         & 90.33 & 63.57 & 52.07  \\
     & TRADES \cite{TRADES}       & 90.38 & 73.31 & 57.94  \\
     & ATA \cite{benchmark_sota}  & 89.11 & 62.81 & 53.75  \\
     & TAAT \cite{benchmark_sota} & 90.44 & 72.59 & 59.91  \\
     & GReAT (SGD)                & 90.24 & \textbf{74.47} & \textbf{63.45}  \\
     & GReAT (ADAM)               & 90.54 & \textbf{75.81} & \textbf{65.66}  \\
    \bottomrule
    \end{tabular}
}
\end{table}

The benchmark Table \ref{results_sota_svhn} displays the robustness outcomes for the SVHN dataset, presenting the accuracy of different methods in countering FGSM and PGD-100 adversarial attacks. The proposed method demonstrates superior performance compared to other approaches in terms of robust accuracy.

\section{Conclusion}
\label{sec:conclusion}

In this paper, we have presented a Graph Regularized Adversarial Training Method (GReAT), designed to enhance the robustness of classifiers. By leveraging classical adversarial training, the graph regularization technique enhances the robustness of deep learning classifiers. This technique employs graph-based constraints to regularize the training process, thereby bolstering the model's capacity to withstand adversarial attacks. Integrating these constraints enables the model to learn more robust features and be less prone to manipulation via adversarial examples. This strategy has demonstrated significant potential to enhance the robustness and generalization of deep learning classifiers, indicating that it is a valuable tool in adversarial training.

\EOD

\bibliography{MyLibrary2, refs_2}

\end{document}